\documentclass[10pt]{article}

\usepackage{times}
\usepackage{amsmath}
\usepackage{amssymb}
\usepackage{cite}
\usepackage{bm}
\usepackage{float}
\usepackage{url}
\usepackage[ruled]{algorithm}
\usepackage{algorithmicx}
\usepackage{algpseudocode}
\usepackage{breqn}
\usepackage{graphicx}
\usepackage{subfigure}
\usepackage[small]{caption}
\usepackage{color}
\usepackage{graphics}
\usepackage{graphicx,epsfig,psfrag,pstricks}
\usepackage{multirow}
\usepackage{array}
\usepackage[round, authoryear]{natbib}

\title{A Parzen-based distance between probability measures as an alternative
of summary statistics in Approximate Bayesian Computation}
\author{Carlos D. Zuluaga, Edgar A. Valencia, Mauricio A. \'Alvarez\\
{\small \emph{Faculty of Engineering, Universidad Tecnol{\'o}gica de Pereira, Colombia, 660003.}}\\}
\date{}

\begin{document}
\maketitle

\begin{abstract}

Approximate Bayesian Computation (ABC) are likelihood-free Monte Carlo
methods. ABC methods use a comparison between simulated data, using
different parameters drew from a prior distribution, and observed
data.  This comparison process is based on computing a distance
between the summary statistics from the simulated data and the
observed data. For complex models, it is usually difficult to define a
methodology for choosing or constructing the summary
statistics. Recently, a nonparametric ABC has been proposed, that uses
a dissimilarity measure between discrete distributions based on
empirical kernel embeddings as an alternative for summary statistics.
The nonparametric ABC outperforms other methods including ABC, kernel
ABC or synthetic likelihood ABC. However, it assumes that the
probability distributions are discrete, and it is not robust when
dealing with few observations.  In this paper, we propose to apply
kernel embeddings using an smoother density estimator or Parzen estimator
for comparing the empirical data distributions, and computing the ABC
posterior. Synthetic data and real data were used to test the Bayesian
inference of our method. We compare our method with respect to 
state-of-the-art methods, and demonstrate that our method is a robust estimator
of the posterior distribution in terms of the number of observations.
\end{abstract}

\section{Introduction}

Many Bayesian applications use inference for nonlinear stochastic
models, where it is expensive or difficult to evaluate the likelihood;
or the normalization constant in Bayesian modelling is also
intractable. Approximate Bayesian Computation (ABC) are
likelihood-free Monte Carlo methods. ABC methods can be employed to
infer posterior distributions without having to evaluate likelihood
functions \citep{Wilkinson08,Toni09}. They simulate data from a model
with different parameter values and compare summary statistics of the
simulated data whit summary statistics of the observed data. There are
many problems associated to how to choose the summary statistics with
the goal of obtaining accepted samples. The choice of these summaries
is frequently not obvious \citep{Joyce08} and in many cases it is
difficult or impossible to construct a general method for finding such
statistics \citep{Fearnhead12}. According to \citet{Park15}, the
selection of the summary statistics is an important stage in ABC
methods that is still an open question.  

Different algorithms for choosing or constructing summary statistics
have been proposed in the literature.  In the state-of-the-art, we can
find linear regression \citep{Fearnhead12}. Another option is to use a
minimum entropy criterion for choosing the summary statistics
\citep{blum13,Nunes10,Blum10}. \citet{Park15} propose a fully
nonparametric ABC that avoids to select manually the summary
statistics, using kernel embeddings \citep{Gretton:ComparisonDistriAAAI:2007}; 
they employ a two-step process:
the first step is to compare the empirical data distributions using
a dissimilarity distance based on Reproducing Kernel Hilbert Spaces
(RKHS). Such a distance is termed Maximum Mean Discrepancy (MMD). In
the second stage, they use another kernel that operates on probability
measures. This nonparametric ABC outperforms other methods, ABC,
kernel ABC or synthetic likelihood ABC, for more details see
\cite{Park15}. However this nonparametric ABC assumes that the
probability distributions are discrete and it is not robust
when there are few observations.

In this paper, we propose a new metric for comparing two data
distributions in a RKHS with application to ABC simulation. The
difference with respect to Park et al. in \citet{Park15} is the
assumption that the probability density functions, for corresponding
probability distributions in RKHS, are continuous probability
functions, estimated using an smoother density estimator or Parzen
estimator. This fact allows us to obtain a biased estimator of the 
dissimilarity distance between two empirical data distributions
that can be written as $\lambda f+(1-\lambda)\text{MMD}$, where $f\in
\mathcal{H}$ ($\mathcal{H}$ is a Hilbert space), and $0\leq\lambda<1$.
It is possible to demonstrate that our estimator presents a 
  lower root mean squared error, between the true parameters and the 
estimated parameter, than
the MMD estimator, for any value of $\lambda$, see
\citet{Muandet13}. The difficulty of calculating dissimilarity metrics
for empirical distributions in RKHS is to compute integrals, these
integrals are usually difficult to solve analytically. In this paper, we
use Gaussian kernels for estimating these probability density
functions.

We compare ABC based in our metric in RKHS with the ABC and K2ABC
proposed by \citet{Park15}. We also show how to use these methods in
combination with sequential Monte Carlo methods.

Experimental results obtained include the application of the different
methods described above over a synthetic dataset and a real
dataset. The synthetic data was created from an uniform mixture
model. We demonstrate that our method is a robust estimator of the
parameter vector in terms of the number of observations. The real data
corresponds to the change of adult blowfly population during a period
of time, as explained by \citet{Wood10}.

\section{Approximate Bayesian Computation based on Kernel Embeddings}

In this section, we briefly expose the different ABC methods. We then
define the metrics based on kernel embeddings employed with the ABC
methods, including the MMD metric and the metric that we propose in
this paper.

\subsection{Short summary on ABC methods}

Approximate Bayesian Computation (ABC) are likelihood-free Monte Carlo
methods. In practical Bayesian models, exact inference may be
intractable due to different reasons: the likelihood
function is expensive or intractable; or the
normalization constant in Bayesian modelling is also intractable. ABC
methods can be employed to infer posterior distributions without
having to evaluate likelihood functions \citep{Wilkinson08,Toni09}. 
These likelihood-free algorithms
simulate data using different parameters drew from prior
distribution and compare summary statistics of the simulated data 
($s\left( {\mathcal {D}'} \right)$)
whit summary statistics of the observed data ($s\left( {\mathcal {D}}
\right)$). 
For this comparison it is necessary to define a tolerance threshold
that determines the accuracy of the algorithm, and a distance measure
$\rho \left( s\left( {\mathcal {D}}
\right),s\left( {\mathcal {D}'} \right) \right)$, for example Euclidean
distance, etc.
If $\rho \left( s\left( {\mathcal {D}}
\right),s\left( {\mathcal {D}'} \right) \right) \leq \epsilon$, 
we then accept the parameters
$\theta$ drew from $p \left( \theta \right)$, otherwise, it is
rejected. 



According to \citet{Wilkinson08}, there are different open questions around
the ABC algorithm including how to choose the measure function $\rho$; what should be the value of
$\epsilon$?; how to select the summary statistics $s$? $\rho$ and
$\epsilon$ are experimental and implementation issues. With respect to $s$,
it is difficult to define a methodology for choosing or constructing 
the summary statistics \citep{Park15}.

\subsection{Metrics between probability measures by using kernels}

The embedding of distributions in a Reproducing Kernel Hilbert Space
(RKHS) is a methodology that allows us to compute distances between
distributions by using kernel functions
\citep{Berlinet:RKHSProbStats:2004}.  According to
\citet{Sriperumbudur10}, a metric $\gamma_k(P, Q)$ over the probability
measures $P$ and $Q$ can be defined through a characteristic kernel\footnote{
  A characteristic kernel is a reproducing kernel for which
  $\gamma_k(P, Q)=0\iff P=Q, P,Q\in\mathcal{P}$, where $\mathcal{P}$ denotes
  the set of all Borel probability measures on a topological space
  $(M,\mathcal{A})$.}
$k(x, x')$ as,
\begin{align}\notag
\gamma_k\left( {P,Q}\right)&=\left\Vert
\int_{M}k\left({\cdot,x}\right)dP\left({x}\right)-\int_{M}k\left({\cdot,y}\right)dQ\left(
    {y}\right)\right\Vert^2_{\mathcal {H}}.
\end{align}
If the distributions $P(x)$, and $Q(y)$ admit a density, then we have
$dP(x) = p(x)dx$, and $dQ(y)=q(y)dy$, and an alternative expression
for $\gamma_k\left( {P,Q}\right)$ can be written as
\begin{dmath}\label{eq26}
  \gamma_k\left( {P,Q}\right) = \int_{M}\int_{M}k\left({x,z}\right)
  p\left({x}\right)q\left({z}\right)dxdz + \int_{M}\int_{M}k
  \left({z,y}\right)p\left({z}\right)q\left({y}\right)dzdy\\
- 2\int_{M}\int_{M}k(x,y)p(x)q(y)dxdy.
\end{dmath}

\subsection{Kernel Embeddings as summary statistics for ABC}

Let us assume that we have two random samples $X = \{x_i\}_{i=1}^{N_x}$, and
$Y = \{y_j\}_{j=1}^{N_y}$. In ABC, one of those samples would correspond to the
real data ($\mathcal{D}$), whereas the other sample would correspond to simulated data ($\mathcal{D}'$) from
the model we wish to estimate. As mentioned before, we need a way to define
if accepting the simulated data. A key idea introduced by \citet{Park15}
was to assume that the random samples $X$, and $Y$ are drawn from probability
measures $P$, and $Q$, respectively, and instead of comparing the distance
between samples $X$, and $Y$, they propose to compare the distance between
the probability measures $P$ and $Q$.

The authors in \citet{Park15} assume empirical distributions for $P$, and $Q$,
this is $p(x) = \frac{1}{N_x}\sum_{i=1}^{N_x} \delta(x-x_i)$, and
$q(y)=\frac{1}{N_y}\sum_{j=1}^{N_y}\delta(y-y_j)$. With these expressions for
$p(x)$, and $q(y)$, the distance $\gamma_k(P,Q)$ is given by
\begin{align}\label{eq22}
\gamma_k\left( {P,Q}\right) &=
\frac{1}{N^2_x}\sum^{N_x}_{i,j=1}k\left({x_i,x_j}\right)
+\frac{1}{N^2_y}\sum^{N_y}_{i,j=1}k\left({y_i,y_j}\right)
-\frac{2}{N_xN_y}\sum^{N_x,N_y}_{i,j=1}k\left({x_i,y_j}\right).
\end{align}
We refer to this distance as $\gamma_{k}^{\text{MMD}}(P,Q)$, since it is
rooted in the Maximum Mean Discrepancy (MMD) concept developed in
\cite{Gretton:kernelTwoSampleNIPS:2007,
  Gretton:kernelTwoSampleJMLR:2012}. After obtaining $\gamma_{k}^{\text{MMD}}(P,Q)$, 
	Park et al. in \citet{Park15} apply a second
Kernel that operates on probability measures, as follows

\begin{align}\label{eq23}
  k_\epsilon\left( {P_{\mathcal {D}},Q_{\mathcal {D}'}}\right)
  & = \exp \left( -\frac{\gamma_k^{\text{MMD}}(P_{\mathcal{D}}, Q_{\mathcal{D}'})}{\varepsilon}
    \right),
\end{align}	

where $\varepsilon$ is a positive parameter for the second kernel. $P_{\mathcal{D}}$
is the distribution associated to $\mathcal{D}$, and $Q_{\mathcal{D}'}$ is
the distribution associated to $\mathcal{D}'$. In \citet{Park15}, the authors use
an unbiased estimate for $\gamma_{k}^{\text{MMD}}(P,Q)$ in which the factors
$1/N_x$, and $1/N_y$ are replaced for $1/(N_x(N_x-1))$, and $1/(N_y(N_y-1))$,
respectively. Also, the innermost sum in each of the first two terms
does not take into account the terms for which $i=j$.

Instead of assuming a discrete distribution for $p(x)$, and $q(y)$, in this
paper we propose to use a smooth estimate for both densities based on the
Parzen-window density estimator. 
We assume that the densities $p(x)$, and $q(x)$ can be estimated using
\begin{align*}
\widehat{p}\left({x}\right)
&=\frac{1}{N_x}\sum_{m=1}^{N_x}\frac{1}{\left({2\pi
    h_p^2}\right)^{D/2}}\exp\left({-\frac{\Vert
    x-x_m\Vert^2}{2h_p^2}}\right),\\
\widehat{q}\left({y}\right)
&=\frac{1}{N_y}\sum_{n=1}^{N_y}\frac{1}{\left({2\pi
    h_q^2}\right)^{D/2}}\exp\left({-\frac{\Vert
    y-y_n\Vert^2}{2h_q^2}}\right),
\end{align*}
where $h_p$ and $h_q$ are the kernel bandwidths, and $D$ is the dimensionality
of the input space.

If we use a Gaussian kernel with parameter
$\Sigma$ for $k(x,x')$, and the estimators $\widehat{p}\left({x}\right)$, and
$\widehat{q}\left({y}\right)$, a new distance
between the distributions $P$ and $Q$ is easily obtained from expression
\eqref{eq22} as follows
\begin{align}\label{eq29}
\gamma_k\left({P, Q}\right)
&=\frac{1}{N_x^2}\sum_{i,j=1}^{N_x}\hat{k}\left(x_i,x_i; 2\Sigma_p\right)
+\frac{1}{N_y^2}\sum_{i,j=1}^{N_y}\hat{k}\left(y_i,y_j; 2\Sigma_q\right)\\
&-\frac{2}{N_xN_y}\sum_{i,j=1}^{N_x, N_y}\hat{k}\left(x_i,y_j; \Sigma_p+\Sigma_q
\right)\notag,
\end{align}
where
\begin{align*}
  \hat{k}\left({x,x';S}\right) &= \frac{|\Sigma|^{1/2}}{|\Sigma + S|^{1/2}}
  \exp\left({-\frac{\left({x-x'}\right)^{\top}\left({\Sigma+S}\right)^{-1}
      \left({x-x'}\right)}{2}}\right).
\end{align*}
In expression \eqref{eq29}, $\Sigma_p = h^2_p\mathbf{I}$ and
$\Sigma_q = h^2_q\mathbf{I}$. We refer to the metric in \eqref{eq29}
as $\gamma_k^{\text{Parzen}}(P,Q)$.\\

As a distance measure $\gamma_k({P_{\mathcal {D}},Q_{\mathcal {D}'}})$
we can use $\gamma_{k}^{\text{MMD}}({P_{\mathcal {D}},Q_{\mathcal
    {D}'}})$ or $\gamma_{k}^{\text{Parzen}}({P_{\mathcal
    {D}},Q_{\mathcal {D}'}})$. The algorithm proposed by \citet{Park15}
that employs kernel embeddings of probability measures in ABC using
MMD is shown in Algorithm \ref{alg:KEABC}. If we use the metric
$\gamma_{k}^{\text{MMD}}({P_{\mathcal {D}},Q_{\mathcal {D}'}})$ on
line 4 for the algorithm, we refer to the method as K2ABC. If we use
the metric $\gamma_{k}^{\text{Parzen}}({P_{\mathcal {D}},Q_{\mathcal
    {D}'}})$ instead, we refer to the method as PABC.

\begin{algorithm}
\caption{ABC based on Kernel Embeddings}\label{alg:KEABC}
\begin{algorithmic}[1]
\State \textbf{Input:} Observed data $\mathcal{D}$, prior distribution, threshold $\varepsilon$. 
  \For{$i = 1,\ldots,N_s$}\\
  Draw $\theta$ from $p \left( \theta \right)$\\
  Simulate $\mathcal {D}'$ using $p\left( {\left. \cdot \right|\theta }
  \right)$ \\
  Compute ${\widetilde w_i}
  = \exp \left( -\frac{\gamma_k(P_{\mathcal{D}}, Q_{\mathcal{D}'})}{\varepsilon}
    \right)$
    \EndFor\\
    Normalize $\widetilde w_i$
\end{algorithmic}
\end{algorithm}

\subsection{Extension to ABC SMC}

The disadvantages of ABC, suck as: the selection of $\epsilon$
, the choice of summary statistics or accepted samples 
with low probability, it can be avoided using an
ABC algorithm based on sequential Monte Carlo methods (ABC SMC)
proposed by \citet{Sisson07}. The goal of ABC SMC is
to obtain an approximation of true posterior using a series of
sequential steps, expressed by $p\left( {\left. \theta \right|{\rho
    \left( {\mathcal {D},\mathcal {D}'} \right) \leq \epsilon_t}}
\right)$, for $i = 1, \cdots , T$, where $\epsilon_t$ is a threshold
that decreases in each step
($\epsilon_1,>,\ldots,\epsilon_t,>,\ldots,>\epsilon_T$), thus it
refine the approximation towards the target distribution.

ABC SMC has a first stage based on ABC rejection. We can replace this
stage with K2ABC or PABC, leading to what we call in the paper as the
K2ABC SMC, and PABC SMC methods. Details of the ABC SMC method can be
found in \citet{Toni09}.

\section{Experimental Setup}\label{section:MAM}

We first describe the synthetic dataset and the real dataset used for
our experiments.  Finally, we explain the procedure for applying the
ABC methods over synthetic and real datasets.

\subsection{Datasets}

We follow the experiments described in \citet{Park15}. Synthetic
data is created from an uniform mixture model. The real dataset correspond to
a time series of an adult blowfly population \citep{Wood10}.\\

\textbf{Toy Problem.} The synthetic dataset was generated by an
uniform mixture model, $p\left( {\left. {\mathcal D} \right|\theta }
\right) = \sum_{k = 1}^K {{\pi _k}{\mathcal U} \left( {k,k -
    1}\right)}$, where $\mathcal {D}$ are the observed data; $\pi _k$
are the mixing coefficients; and $K$ is the number of components. The
model parameters $\boldsymbol{\theta}$ correspond to the mixing
coefficients $\boldsymbol{\pi} = {\left[ {0.25,0.04,0.33,0.04,0.34}
    \right]^\top}$. The prior distribution for $\boldsymbol{\theta}$
is a symmetric Dirichlet distribution.\\

\textbf{Noisy nonlinear ecological dynamic system} For the real
dataset example, we would like to infer the parameters of a non-linear
ecological dynamic system \citep{Meeds14} represented through ${N_{t +
    1}} = P{N_{t - \tau }}\exp \left( { - \frac{{{N_{t -\tau
  }}}}{{{N_0}}}} \right){e_t} + {N_t}\exp \left({\partial {\varepsilon
    _t}} \right)$, where $N_t$ is the adult population at time $t$,
and $P$, $N_0$, $\partial$ and $\tau$ are parameters. Variables $e_t$ and $\varepsilon _t$ are employed
to represent noise, and they are assumed to follow Gamma distributions $e_t \sim 
{\mathcal G}\left(1/ \sigma_p^2, 1/\sigma _p^2\right)$, and $\varepsilon_t \sim {\mathcal G}
\left(1/\sigma_d^2,1/\sigma_d^2\right)$.  The parameter vector we want to infer is given by
$\boldsymbol{\theta} = {\left[{P,N_0,\partial,\tau,\sigma_p,\sigma_d} \right]^\top}$. The observed
data that we use in the experiments correspond to a time series of the adult population of sheep blowfly, with
$180$ observations. The data was provided by the authors of \citet{Wood10}.

\subsection{Validation}

We perform a comparison between ABC, K2ABC, PABC, SMCABC, K2ABC SMC,
and PABC SMC.  The comparison is made in terms of the
root-mean-squared error (RMSE) between the true parameters (when
known) and the estimated parameters by the different methods. For the
real dataset, we compute the cross-correlation coefficient ($\rho$)
between the real time series and the time series generated by the
different simulation methods.

\subsection{Procedure}\label{subsecion:Procedure}

For the toy problem, the observed data $\mathcal{D}$ is sample once
from the mixture of uniform densities with the mixture coefficients
$\bm{\theta}$ described above. The observed data contains $400$
observations. To generate the simulated data $\mathcal{D}'$, we sample
from the Dirichlet prior, and with that sample, we then generate $400$
observations from the mixture model. We then apply the ABC, K2ABC,
PABC, ABC SMC, K2ABC SMC and PABC SMC to the observed data, and the
simulated data, and compute the RMSE over the true and estimated
parameters. For all the ABC methods, the procedure for generating
simulated data is repeated $1000$ times. For ABC, we use $\epsilon =
0.002$, and compute the mean and standard deviation as summary
statistics. For ABC SMC, K2ABC SMC and PABC SMC, we employ
$\left\{ {{\epsilon_t}} \right\}_{t = 1}^T = \left( {0.5,0.01,0.005,0.001,0.0005}\right)$. 
For the ABC SMC type of algorithms, we need to specify what
is known as a perturbation kernel. For this, we use a spherical
multivariate Gaussian distribution, with parameter $0.0001$.

For the noisy nonlinear ecological dynamic system, we apply all the
ABC methods and calculate the cross-correlation coefficient between
the observed data ($\mathcal {D}$), and the simulated data ($\mathcal
{D}'$), using each method. We adopt similar priors to the ones used by
\citet{Meeds14}: $\log P \sim \mathcal {N}\left({3,0.2}\right)$, $\log
\partial \sim \mathcal {N}\left({-1.5,0.1}\right)$,
$\log N_0 \sim \mathcal {N}\left({6,0.2}\right)$,
$\log \sigma_d \sim \mathcal{N}\left({-0.1,0.01}\right)$,
$\log \sigma_p \sim \mathcal {N}\left({0.1,0.01}\right)$
and $\tau \sim \mathcal {P}\left({6}\right)$.

For ABC, we use $\epsilon = 0.35$, and compute $8$ summary statistics
used in \citet{Meeds14}: $4$ statistics using the log of the mean of
all $25\%$ quantiles of $\left\{ {\frac{{{N_t}}}{{1000}}} \right\}_{t
  = 1}^{180}$, $4$ statistics employing the mean of $25\%$ quantiles
of the first-order differences of $\left\{ {\frac{{{N_t}}}{{1000}}}
\right\}_{t = 1}^{180}$, and we also compute the maximum and minimum
value of $\left\{ {\frac{{{N_t}}}{{1000}}} \right\}_{t =
  1}^{180}$. For ABC SMC, K2ABC SMC and PABC SMC, we employ
$\left\{ {{\epsilon_t}} \right\}_{t = 1}^T = \left( {2,1,0.5,0.35,0.25,0.2,0.15}\right)$; 
we also specify what is known as a perturbation kernel. For this, we use a spherical
multivariate Gaussian distribution, with parameter $0.0001$.

For K2ABC and PABC, in both experiments, we use a kernel bandwidth optimization approach
based on minimizing the mean integrated square error (MISE) between
the observed data and simulated data, to define the values of
$\Sigma$, $h_x$ and $h_y$. For more details about this kernel
bandwidth optimization approach, see \citet{Shimazaki10}.

\section{Results and discussion}\label{section:RAD}

The ABC, K2ABC, PABC, ABC SMC, K2ABC SMC, PABC SMC were evaluated and
compared over the synthetic data set and the real data set described
in section \ref{section:MAM}.

\subsection{Results from synthetic data}

\begin{figure}[htbp]
     \begin{center}
     \psfrag{Mean}[c][][0.45]{{Mean}}
 		 \psfrag{theta}[c][][0.45]{{$\theta$}}
        \subfigure[ABC]{%
            \label{fig:ABC}
            \includegraphics[width=0.3\textwidth]{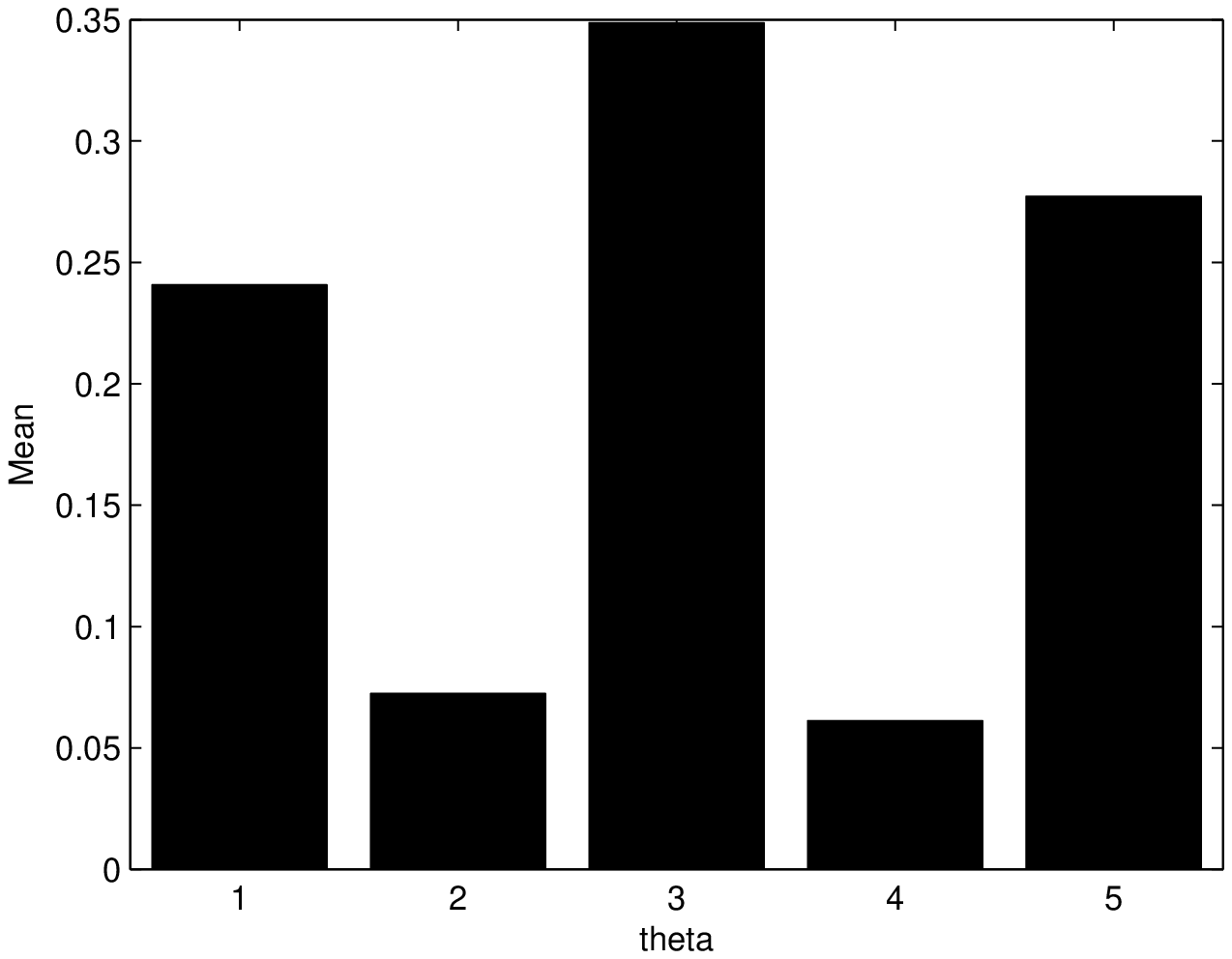}
        }%
        \subfigure[K2ABC]{%
           \label{fig:K2ABC}
           \includegraphics[width=0.3\textwidth]{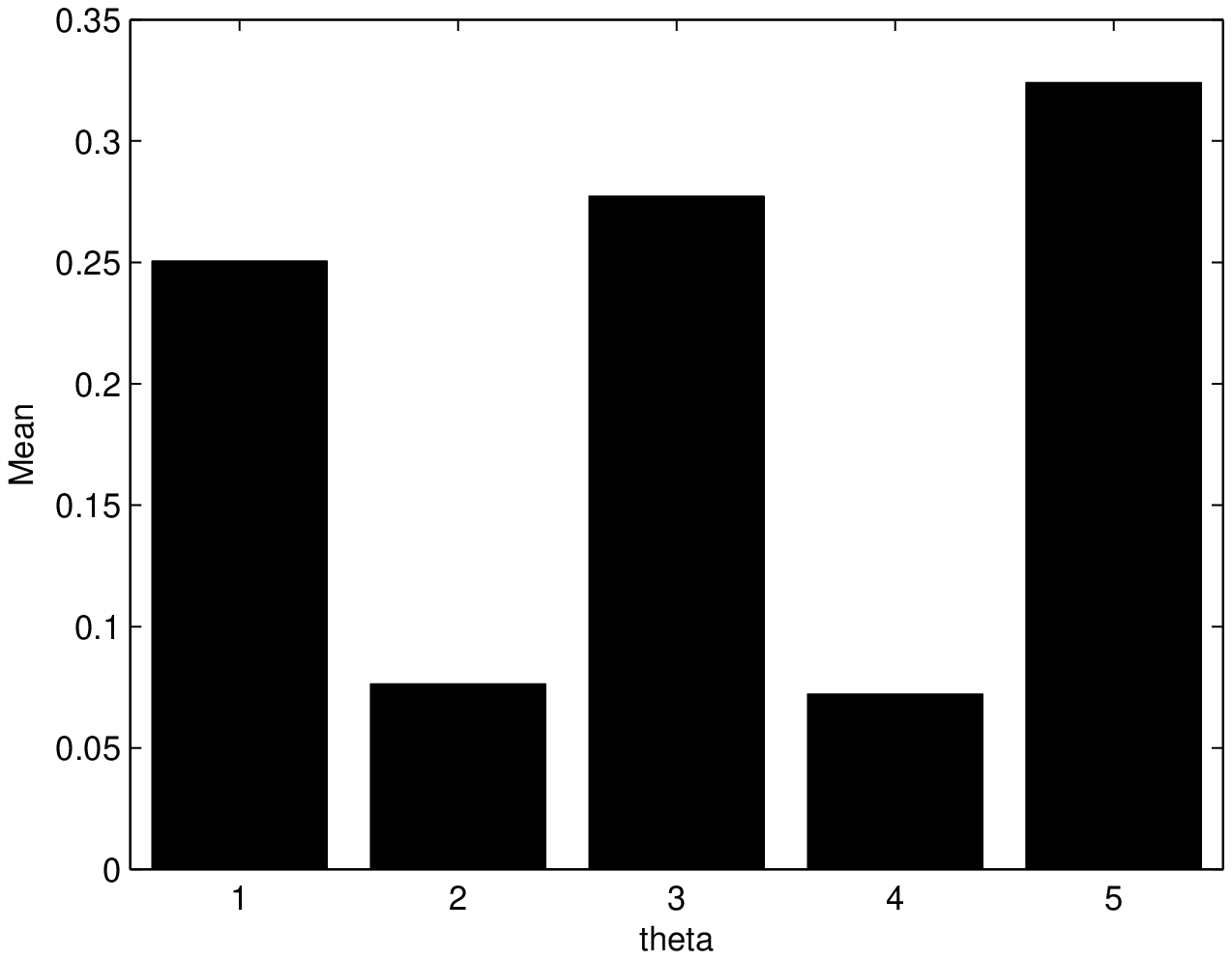}
        }
				\subfigure[PABC]{%
           \label{fig:PABC}
           \includegraphics[width=0.3\textwidth]{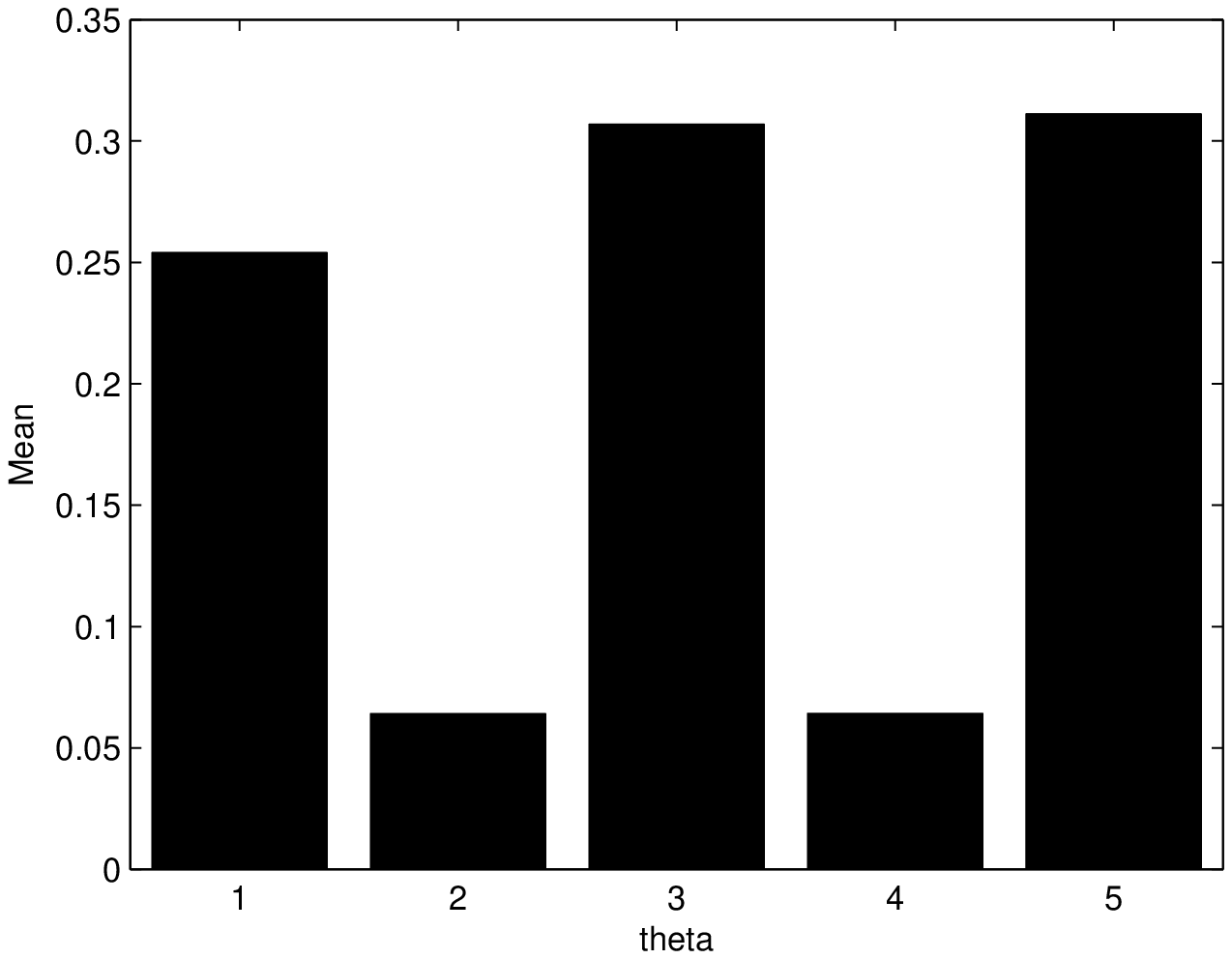}
        }\\ 
        \subfigure[ABC SMC]{%
            \label{fig:ABCSMC}
            \includegraphics[width=0.3\textwidth]{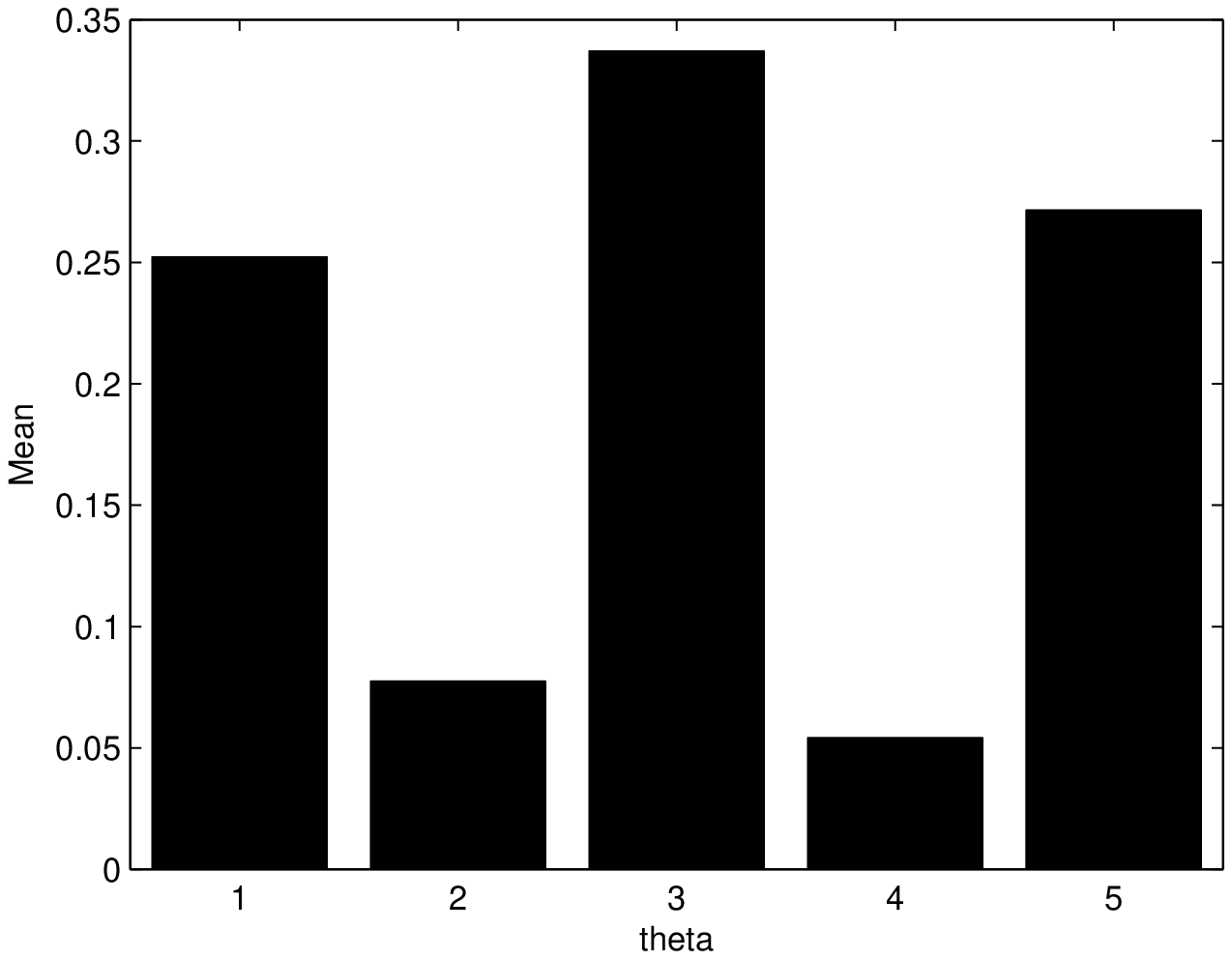}
        }%
        \subfigure[K2ABC SMC]{%
            \label{fig:K2ABCSMC}
            \includegraphics[width=0.3\textwidth]{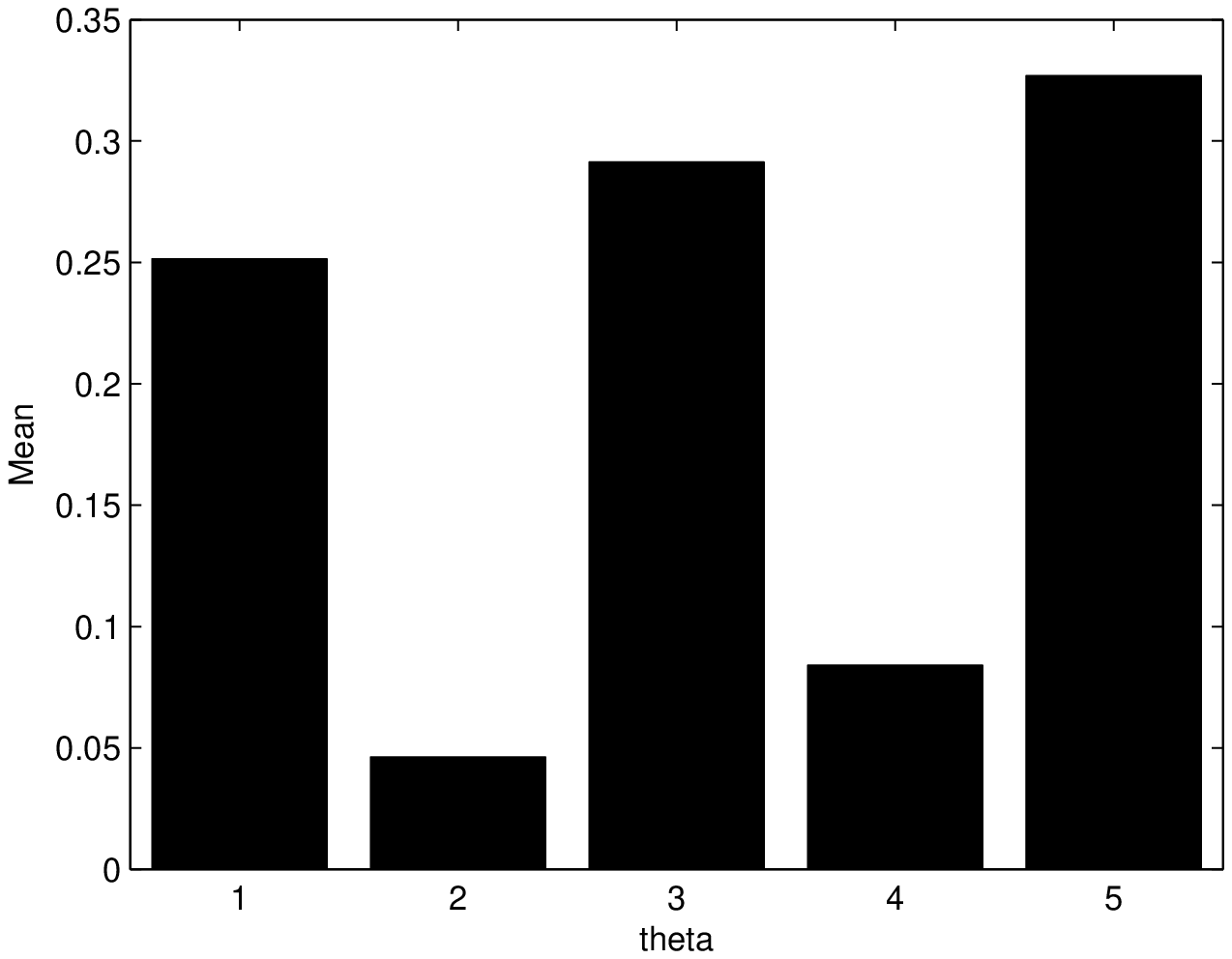}
        }%
				\subfigure[PABC SMC]{%
           \label{fig:PABCSMC}
           \includegraphics[width=0.3\textwidth]{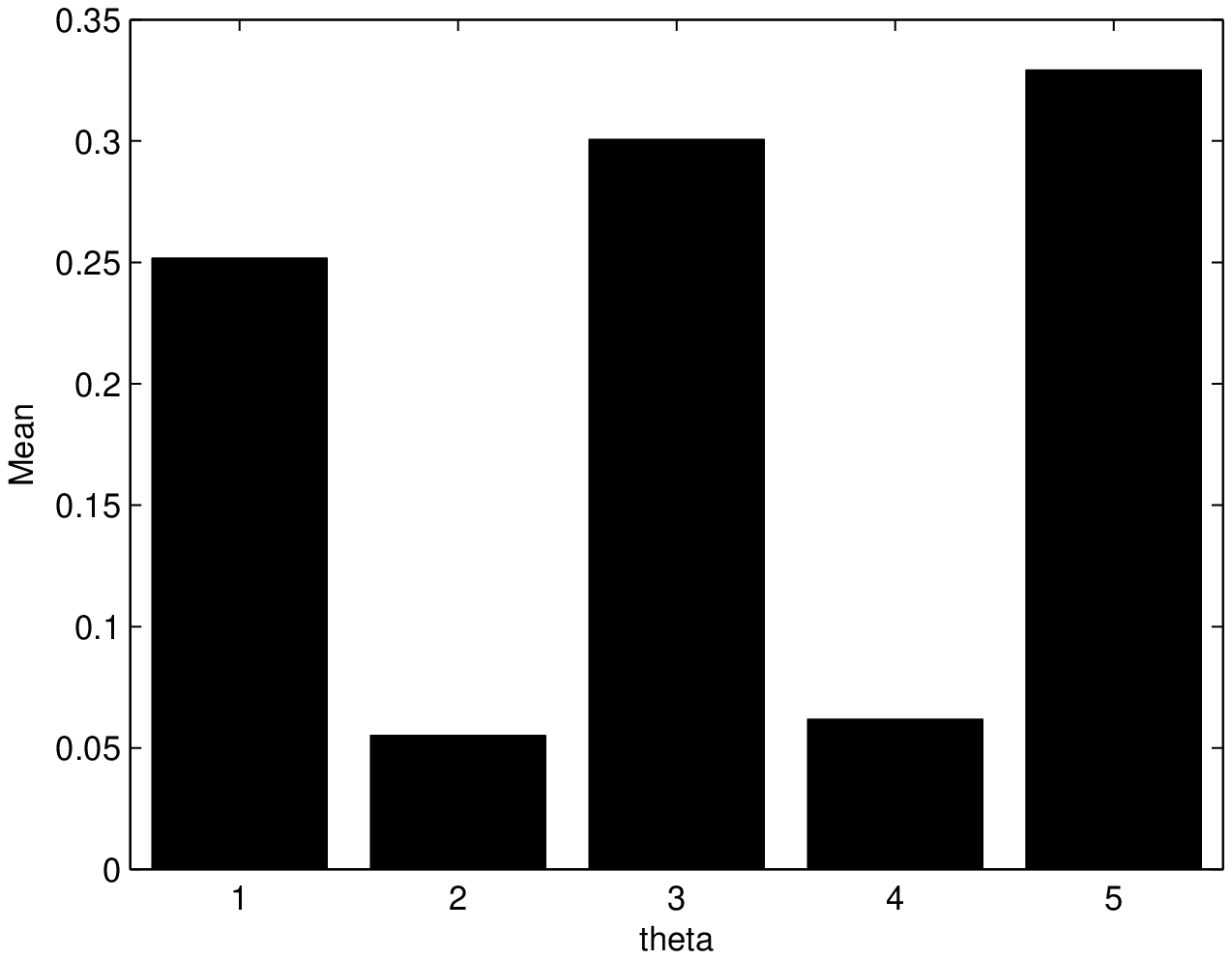}
        }
    \end{center}
    \caption{Estimated posterior mean of model parameters for the
      uniform mixture model, using all ABC methods. The integer number
      in the $x$ axis represents the subindex $i$ in the parameter
      $\theta_i$.}%
   \label{fig:ToyProABCs}
\end{figure}

All ABC methods were applied over vectors of $400$ observations obtained 
from the uniform mixture model described in section
\ref{section:MAM}. Fig. \ref{fig:ToyProABCs} shows a comparison among
the evaluated methods. The figure presents the estimated
$\mathbb{E}\left[ {\left. \theta \right|\mathcal {D}'} \right]$ for
each method.

\begin{figure}[htbp]
     \begin{center}
     \psfrag{RMSE}[c][][0.5]{{RMSE}}
 		 \psfrag{Number of the observations}[c][][0.5]{{Number of observations}}
		\psfrag{ABC}[c][][0.4]{{ABC}}
		\psfrag{K2ABC}[c][][0.4]{{K2ABC}}
		\psfrag{PABC}[c][][0.4]{{PABC}}
		\psfrag{ABCSMC}[c][][0.4]{{ABCSMC}}
		\psfrag{K2ABCSMC}[c][][0.4]{{K2ABCSMC}}
		\psfrag{PABCSMC}[c][][0.4]{{PABCSMC}}
        \subfigure[RMSE for ABC methods]{%
            \label{fig:RMSEABC}
            \includegraphics[width=0.4\textwidth]{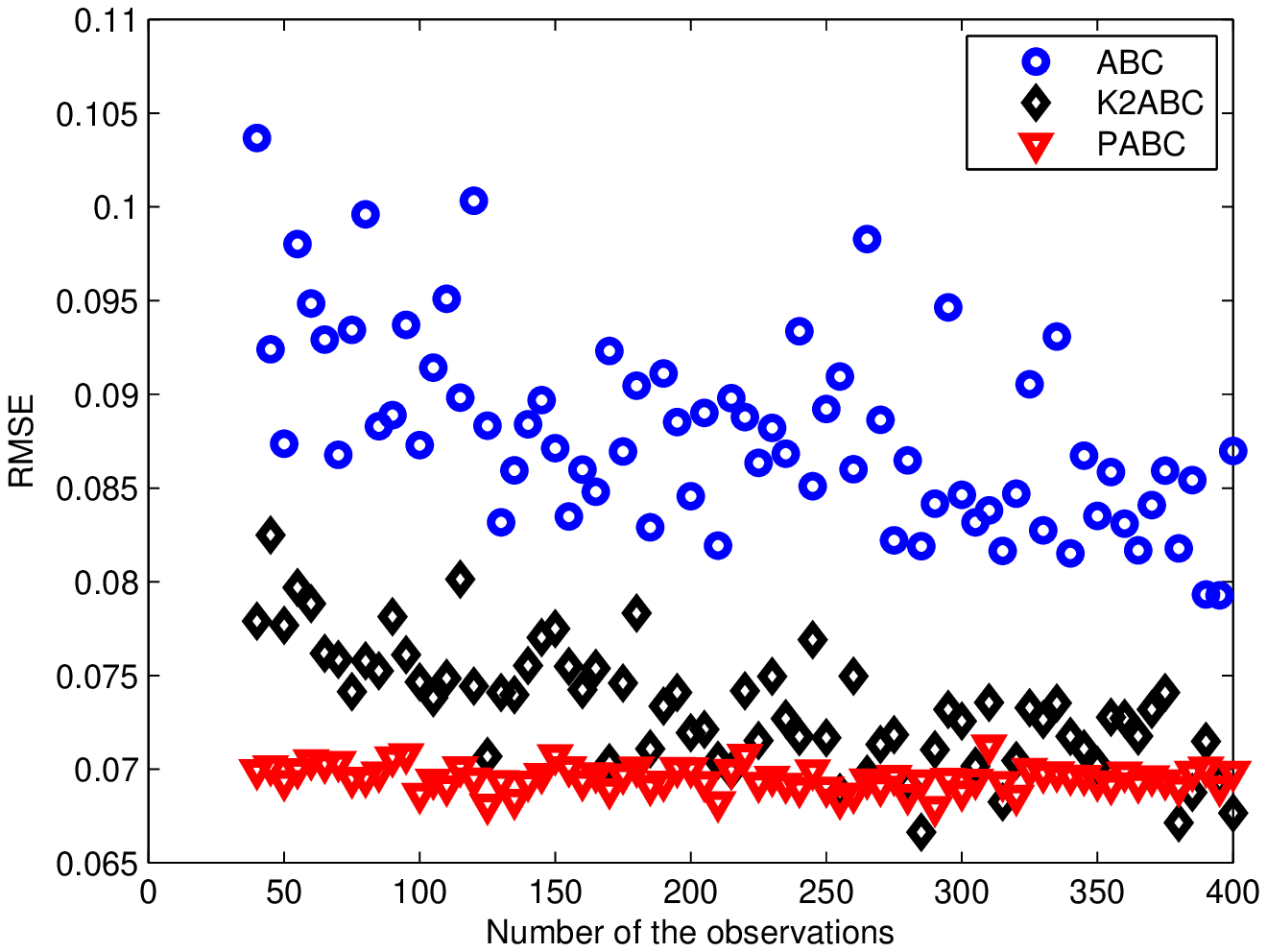}
        }%
        \subfigure[RMSE of ABC methods with SMC]{%
           \label{fig:RMSESMC}
           \includegraphics[width=0.4\textwidth]{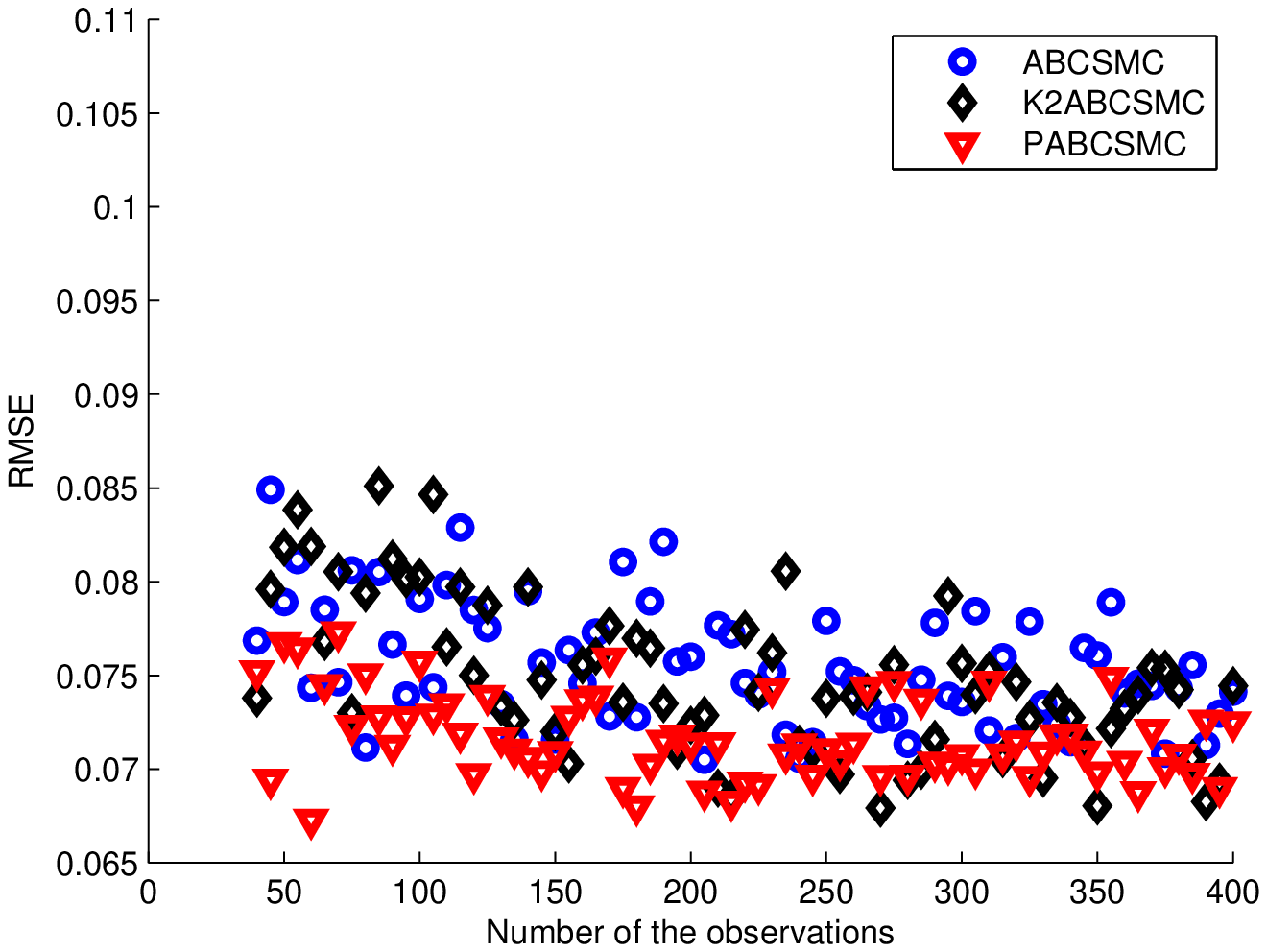}
        }
    \end{center}
    \caption{%
      Root-mean-square error for different number of observations. We start to 
increase the observations from $40$ to $400$, in steps of $5$ observations. 
In Fig. \ref{fig:RMSEABC}, the circles represent the RMSE obtained by ABC, 
the diamonds are the RMSE for K2ABC and the triangles are the RMSE values using 
PABC. In Fig. \ref{fig:RMSESMC}, the circles represent the RMSE obtained by 
ABC SMC, the diamonds are the RMSE for K2ABC SMC and the triangles are the RMSE 
values using PABC SMC.}%
   \label{fig:ToyProRMSEs}
\end{figure}

For K2ABC, PABC, K2ABC SMC and PABC SMC the estimated parameters are
close to the true posterior mean of the parameters (see section
\ref{section:MAM} for the true parameters). ABC and ABC SMC do not 
correctly estimate to $\theta_3$ and $\theta_5$.
To observe the quality of the prediction using each method, we varied
the number of observations and compute the RMSE between the true
parameter vector and the estimated posterior mean of the parameter
vector. We increase the observations from $40$ to $400$, in
steps of $5$ observations.  For each step, we ran all methods and
computed the RMSE. Fig. \ref{fig:ToyProRMSEs} presents the RMSE when
the number of observations is increased.

Comparing Figs. \ref{fig:RMSEABC} and \ref{fig:RMSESMC} show that the
RMSE obtained by PABC is the steadiest and present the lowest
variability, indicating a robust estimator of the parameter vector in
terms of the number of observations. The mean and one standard
deviations for the RMSE for PABC was $0.0696 \pm 0.0006$. The PABC SMC
also present a low variability for this metric, the RMSE value was
$0.0716 \pm 0.0022$. RMSE
obtained by ABC was $0.0879 \pm 0.0050$, for K2ABC was $0.0733 \pm
0.0031$, for ABCSMC was $0.0755 \pm 0.0032$ and for K2ABCSMC was
$0.0747 \pm 0.0041$. For ABC, K2ABC, ABC SMC and K2ABC SMC, notice how
the RMSE decreases when the number of observations increases.

\subsection{Results from nonlinear ecological dynamic system}

\begin{figure}[htbp]
     \begin{center}
     \psfrag{Density}[c][][0.5]{{Density}}
 		 \psfrag{P}[c][][0.5]{{$P$}}
		\psfrag{N0}[c][][0.5]{{$N_0$}}
		\psfrag{sigmad}[c][][0.5]{{$\sigma_d$}}
		\psfrag{sigmap}[c][][0.5]{{$\sigma_p$}}
		\psfrag{delta}[c][][0.5]{{$\partial$}}
		\psfrag{tau}[c][][0.5]{{$\tau$}}
        \subfigure[P]{%
            \label{fig:DistP}
            \includegraphics[width=0.3\textwidth]{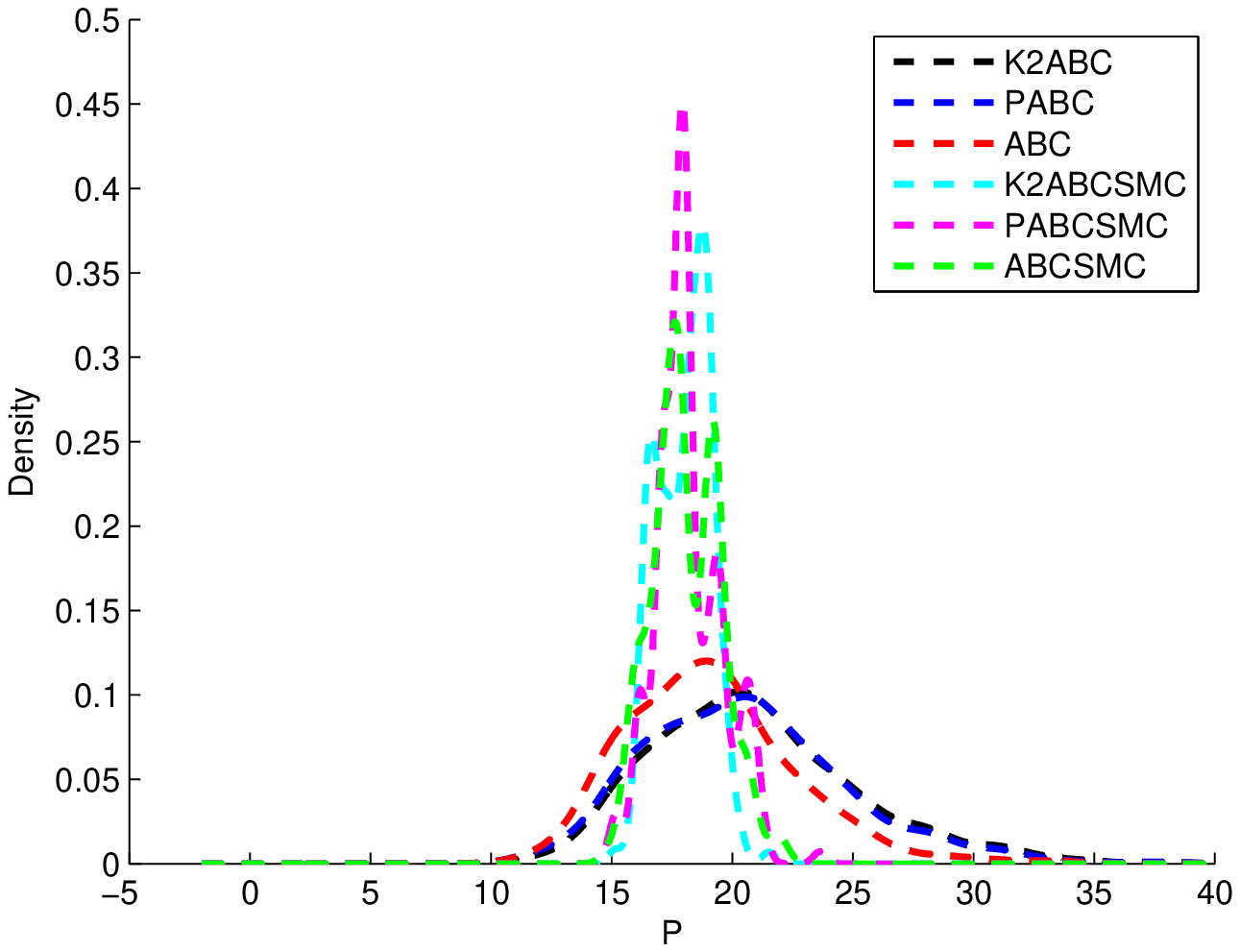}
        }%
        \subfigure[$\partial$]{%
           \label{fig:DistPartial}
           \includegraphics[width=0.3\textwidth]{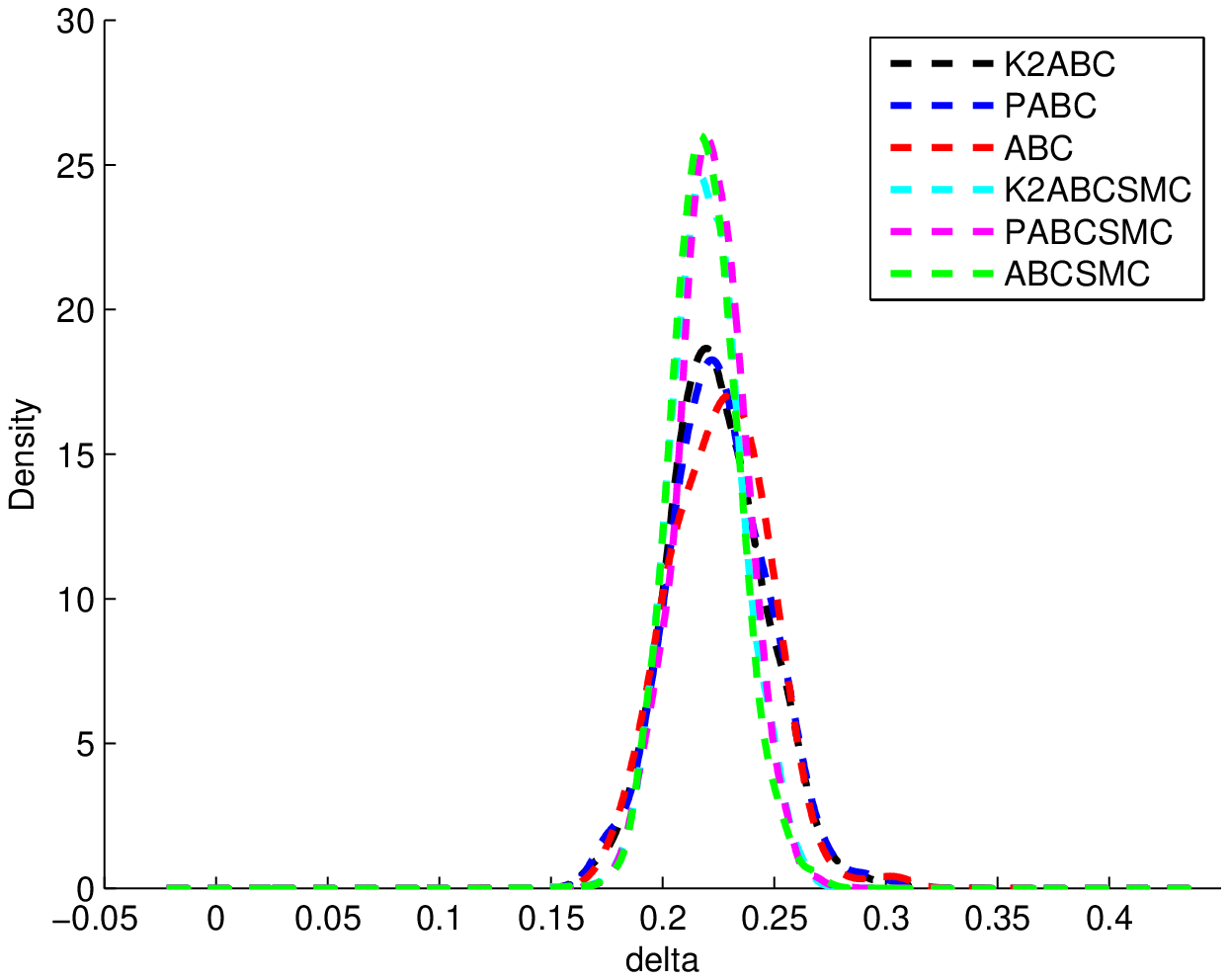}
        }
				\subfigure[$N_0$]{%
           \label{fig:DistN0}
           \includegraphics[width=0.3\textwidth]{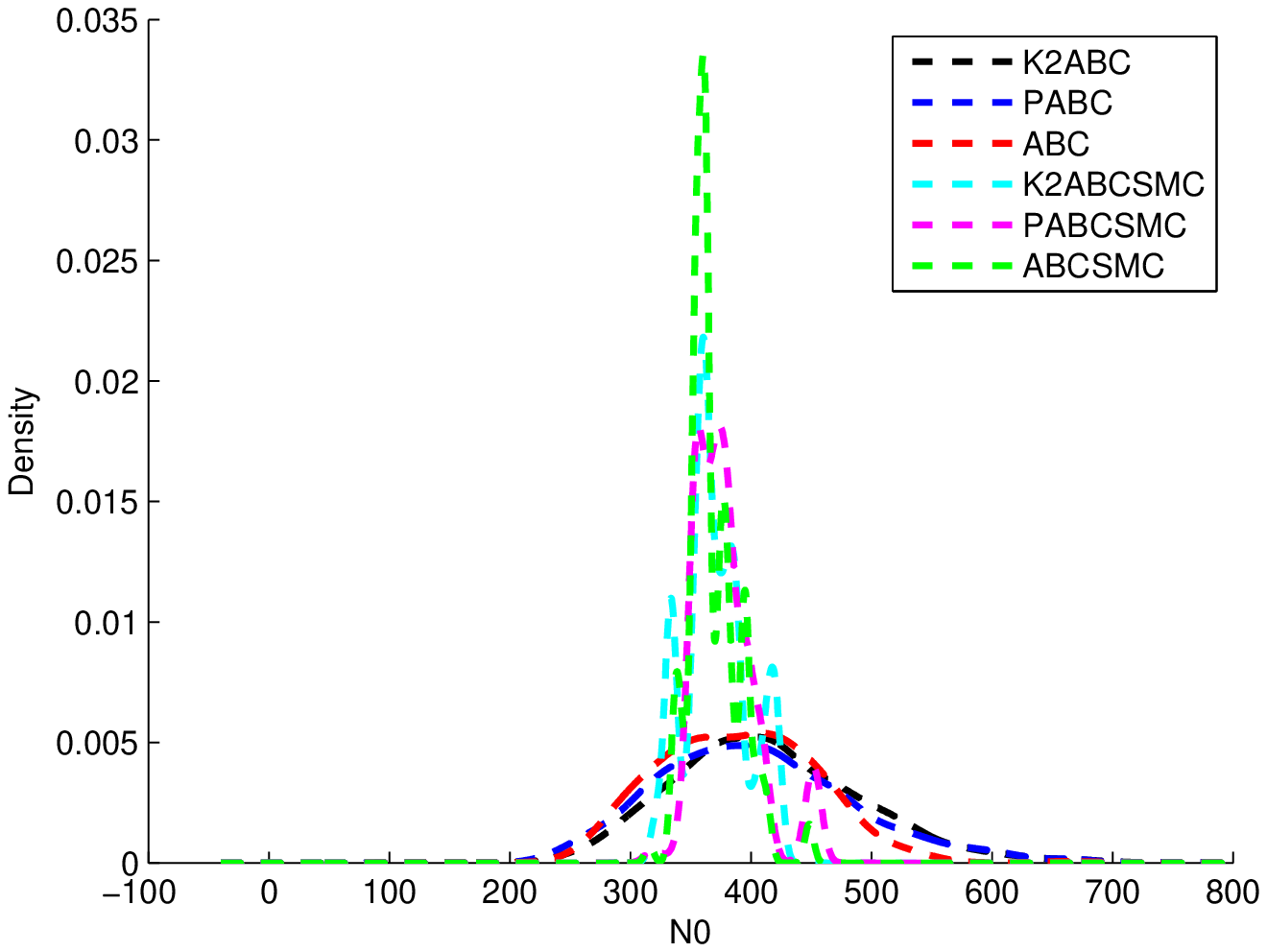}
        }\\ 
        \subfigure[$\sigma_d$]{%
            \label{fig:DistSigd}
            \includegraphics[width=0.3\textwidth]{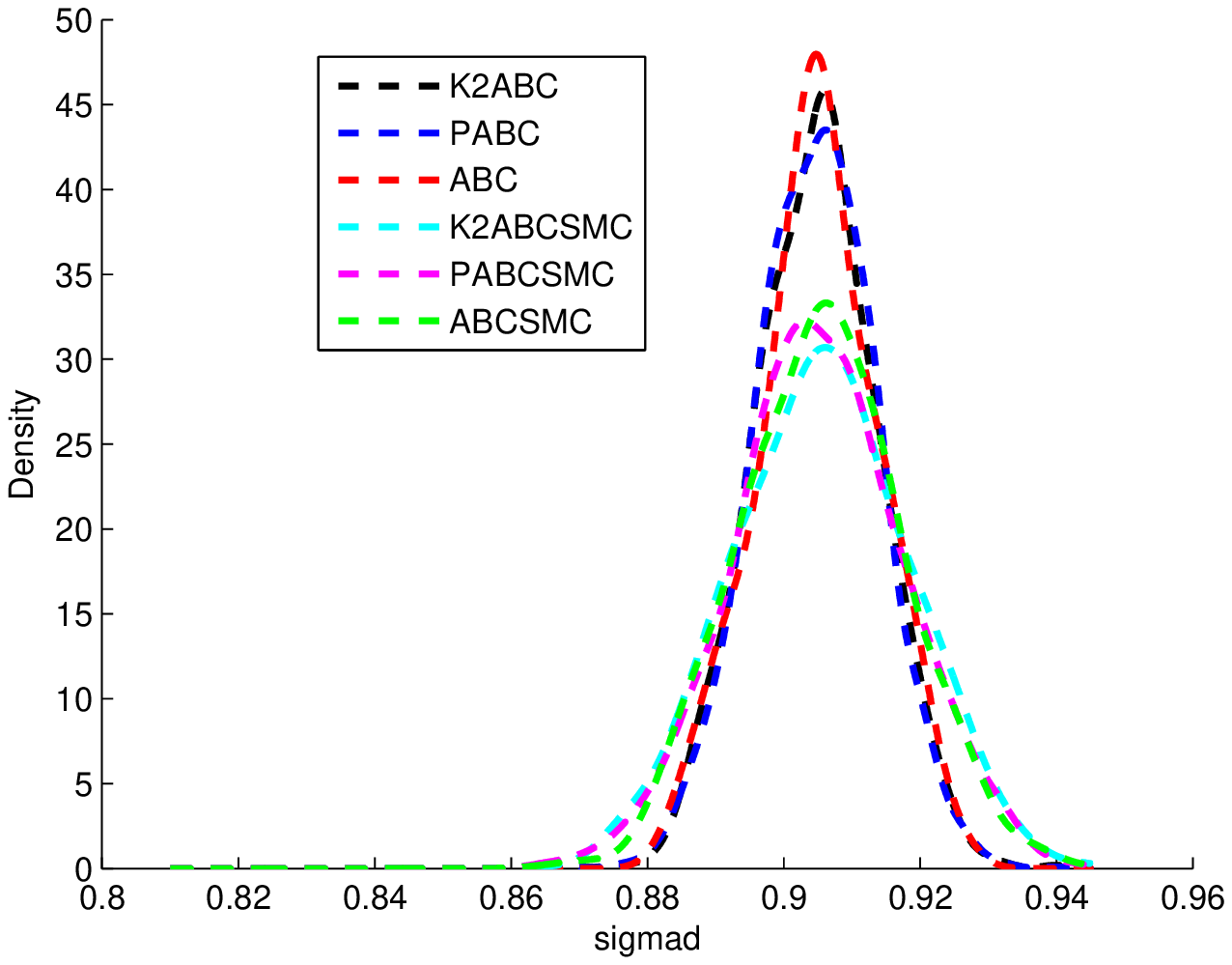}
        }%
        \subfigure[$\sigma_p$]{%
            \label{fig:DistSigp}
            \includegraphics[width=0.3\textwidth]{figures/DistSigmad.eps}
        }%
				\subfigure[$\tau$]{%
           \label{fig:DistTau}
           \includegraphics[width=0.3\textwidth]{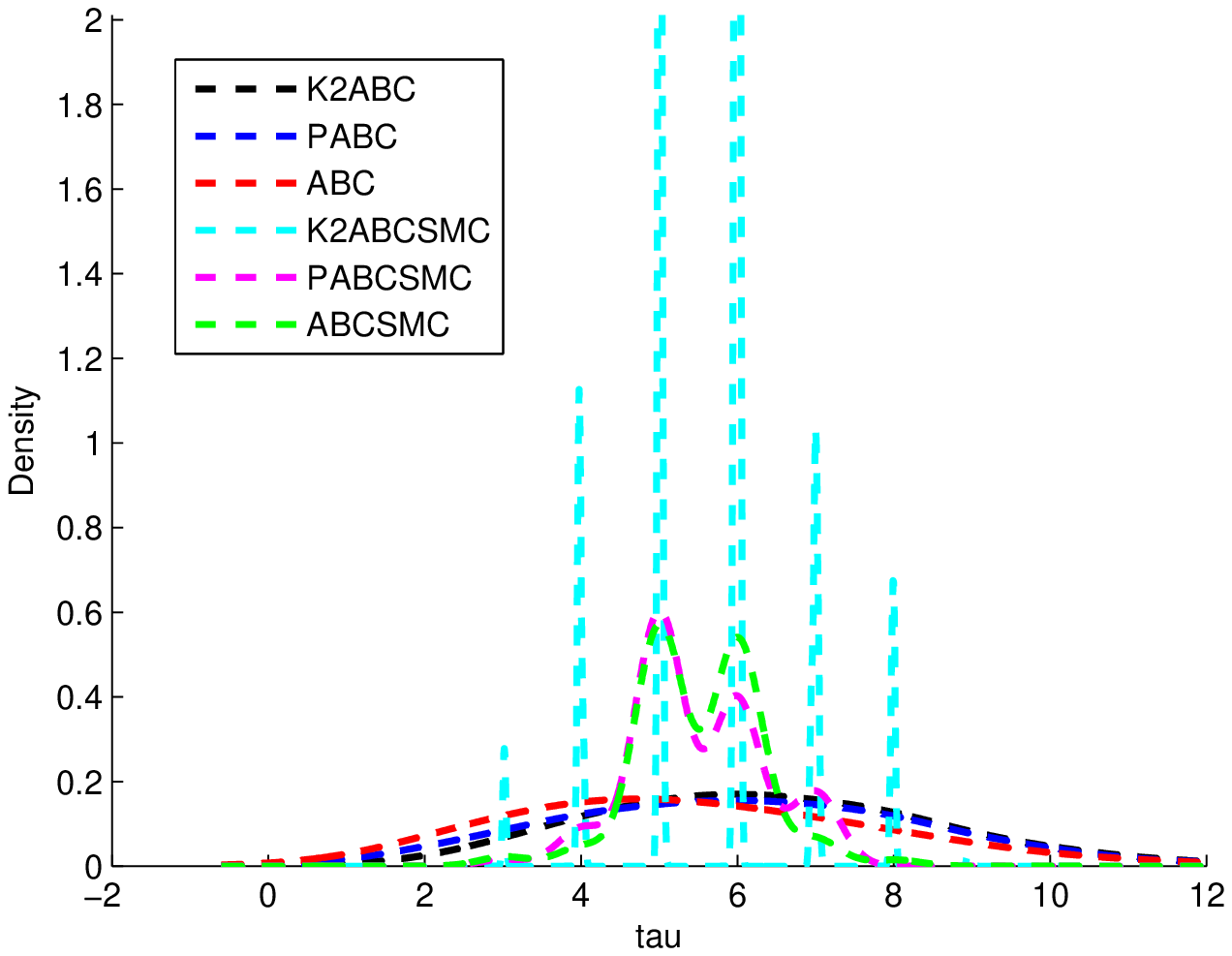}
        }
    \end{center}
    \caption{%
      Posterior distribution for parameters of the nonlinear ecological dynamic system using ABC, K2ABC, PABC, ABC SMC, K2ABC SMC and PABC SMC.}%
   \label{fig:DistEcoDynSyt}
\end{figure}

In Fig. \ref{fig:DistEcoDynSyt}, we present the posterior distribution for the
parameters, obtained for the different methods when applied to the blowfly
dataset of section \ref{section:MAM}. Figs. \ref{fig:DistP},
\ref{fig:DistPartial}, \ref{fig:DistN0}, \ref{fig:DistSigd},
\ref{fig:DistSigp} and \ref{fig:DistTau} illustrate the posterior of
$P$, $\partial$, $N_0$, $\sigma_d$. $\sigma_p$ and $\tau$,
respectively. In these figures, the black dashed line corresponds to
the posterior of the each parameter using K2ABC, the blue dashed line
is the posterior obtained by PABC, the red dashed line is the
posterior employing ABC, the posterior obtained by K2ABC SMC is the
cyan dashed line, the magenta dashed line is the posterior using PABC
SMC and the green dashed line is the posterior obtained by ABC SMC.
The posterior distributions obtained by using all ABC methods for
$\partial$, $\sigma_d$. $\sigma_p$ and $N_0$ are similar. For $P$ and
$\tau$, the posterior obtained by K2ABC SMC, PABC SMC and ABC SMC are
not similar with respect to the posterior using K2ABC, PABC and ABC;
it is due to the parameter values, since
this nonlinear model has a
specific dynamical range from stable equilibrium to chaos.

\begin{figure}[htbp]
	\centering
	\psfrag{ABC}[c][][0.5]{{ABC}}
		\psfrag{K2ABC}[c][][0.5]{{K2ABC}}
		\psfrag{PABC}[c][][0.5]{{PABC}}
		\psfrag{ABCSMC}[c][][0.5]{{ABCSMC}}
		\psfrag{K2ABCSMC}[c][][0.5]{{K2ABCSMC}}
		\psfrag{PABCSMC}[c][][0.5]{{PABCSMC}}
		\psfrag{rho}[c][][0.65]{{$\rho$}}
		\includegraphics[width=0.6\textwidth]{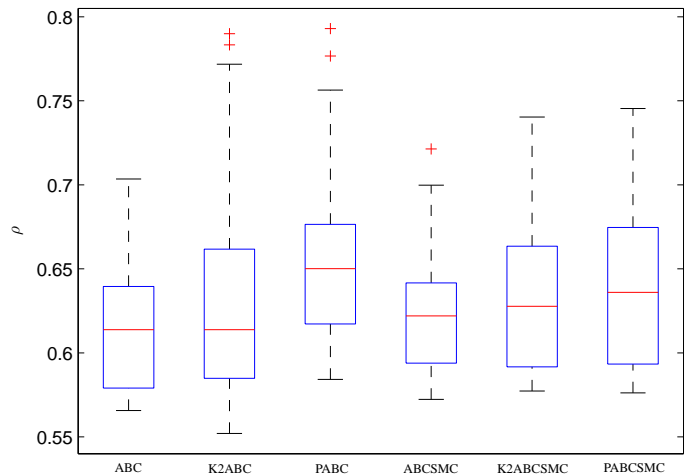}
	\caption{Boxplots of the $25$th and $75$th percentiles of the
          cross-correlation coefficient, using all ABC methods. We drew $100$
subsets of parameters, we then apply all ABC methods, obtaining $100$ 
estimated posterior mean for the parameters, with these estimated parameters
we obtained $100$ time series for these
sets of signals, we compute $\rho$, we then sort $\rho$ in descending
order and choose the $50$ first values, for all
methods.}
	\label{fig:BoxPotCorrCoe}
\end{figure}

To quantify the performance of all methods over the estimated
parameters, we compare the time series obtained using the posterior
means of the parameter and the observed data. For this comparison, we
use a cross-correlation coefficient ($\rho$) between simulated data
$\mathcal{D}'$ and observed data $\mathcal {D}$. The cross-correlation
coefficient measures the similarity between two signals. We drew $100$
subsets of parameters, we then apply all ABC methods, obtaining $100$ 
estimated posterior mean for the parameters, with these estimated parameters
we obtained $100$ time series for these
sets of signals, we compute $\rho$, we then sort $\rho$ in descending
order and choose the $50$ first values, for all
methods. Fig. \ref{fig:BoxPotCorrCoe} contains a box plot for the
cross-correlation coefficient ($\rho$) using the different methods.

From Fig. \ref{fig:BoxPotCorrCoe}, notice that PABC presents the
highest median for $\rho$, with a value of $0.6501$. We also
observe that PABC SMC obtained a median of $0.6360$ for $\rho$. K2AB
CSMC presents a median of $0.6277$, and for ABC SMC the obtained
median for $\rho$ was $0.6220$. Finally, K2ABC and ABC obtained a
median of $0.6138$. 

\section{Conclusions}\label{section:Conclusions}

We introduced a new metric for comparing two data distributions in a
RKHS, using smoother density estimators to compare empirical data
distributions, and then highlight the accepted samples by employing
ABC methods. We demonstrated that our method is a robust estimator of
the parameter vector in terms of the number of observations. Finally,
we showed for a real application that our method obtained the best similarity
with respect to the observed data, in an application involving time-series.
As future work, it would be possible to propose a new dissimilarity
distance using RKHS for different applications like
electrical networks analysis.

\subsubsection*{Acknowledgements} 

We thank the authors of \citet{Meeds14} who kindly sent us the blowfly
population database. C. D. Zuluaga is being funded by the Department
of Science, Technology and Innovation, Colciencias. E. A. Valencia is
being partly funded by Universidad Tecnol\'ogica de
Pereira. M. A. \'Alvarez would like to thank to Colciencias and
British Council for funding under the research project ``Hilbert Space
Embeddings of Autoregressive Processes''.

\bibliographystyle{plainnat}
\bibliography{pbib}

\end{document}